\documentclass{article}
\usepackage{iclr2027_conference,times}

\usepackage{amsmath,amsfonts,bm}

\def\eqref#1{equation~\ref{#1}}

\def\1{\bm{1}}

\DeclareMathAlphabet{\mathsfit}{\encodingdefault}{\sfdefault}{m}{sl}
\SetMathAlphabet{\mathsfit}{bold}{\encodingdefault}{\sfdefault}{bx}{n}

\usepackage{hyperref}
\usepackage{url}
\usepackage{booktabs}
\usepackage{amsmath}
\usepackage{amssymb}
\usepackage{graphicx}
\usepackage{xcolor}
\usepackage{multirow}
\usepackage{enumitem}
\usepackage{float}
\usepackage{tikz}
\usetikzlibrary{positioning,fit,calc}
\usepackage{listings}
\definecolor{actc}{HTML}{D55E00}
\definecolor{absc}{HTML}{0072B2}
\definecolor{inkc}{HTML}{1A1F24}
\definecolor{mutedc}{HTML}{6B7480}
\definecolor{boxbg}{HTML}{F4F6F8}
\definecolor{darkblue}{rgb}{0, 0, 0.5}
\hypersetup{pdftitle={The Model Says Walk: Measuring Whether LLMs Condition on Hidden Constraints}, pdfauthor={Yubo Li, Lu Zhang, Tianchong Jiang, Ramayya Krishnan, Rema Padman}, colorlinks=true, citecolor=darkblue, linkcolor=darkblue, urlcolor=darkblue}

\newcommand{\pA}{p_{\mathrm{A}}}
\newcommand{\pR}{p_{\mathrm{R}}}
\newcommand{\Sel}{S}
\newcommand{\SBA}{\mathrm{SBA}}
\newcommand{\OA}{\mathrm{OA}}
\newcommand{\UA}{\mathrm{UA}}
\newcommand{\core}{\textsc{Core-100}}
\newcommand{\hob}{\textsc{HOB}}
\newcommand{\cond}[1]{\textsc{#1}}
\newcommand{\act}{C}
\newcommand{\heu}{H}

\title{The Model Says Walk: Measuring Whether\\ LLMs Condition on Hidden Constraints}

\author{
Yubo Li$^{1}$\thanks{Corresponding author: \texttt{yubol@andrew.cmu.edu}} \quad
Lu Zhang$^{2}$ \quad
Tianchong Jiang$^{2}$ \quad
Ramayya Krishnan$^{1}$ \quad
Rema Padman$^{1}$ \\[0.5ex]
$^{1}$Carnegie Mellon University \quad
$^{2}$Independent Researcher
}

\newcommand{\codeurl}{\textcolor{red}{\texttt{[CODE-URL]}}}

\iclrfinalcopy
\begin{document}

\maketitle
\lhead{Preprint.}

\begin{abstract}
Asked whether to walk or drive to a car wash 50\,m away, most language models say \emph{walk}, forgetting that the car has to be there. Such failures are usually measured by accuracy on questions where the hidden constraint applies. We show that this is misleading: a model can answer these questions correctly without reasoning about the constraint at all, simply by favoring the cautious option. We instead ask whether a model's decision changes when the constraint is removed. We test this at three levels of control: log-probability sweeps on open-weight models, a 500-prompt stress test, and \core{}, a new human-validated benchmark of minimal constraint-present/constraint-absent pairs. The picture is consistent. The constraint nudges decisions rather than governing them. Models still miss presence constraints like the car wash, yet elsewhere they apply constraints that are not there. As a result, standard accuracy flatters all ten models we evaluate and reorders their ranking, and prompting fixes that look effective largely vanish under paired scoring. Claims about hidden-constraint reasoning need paired evidence.
\end{abstract}

\section{Introduction}
\label{sec:intro}

In February 2026 a one-line question spread across social media: \emph{``I want to wash my car. The car wash is 50 meters away. Should I walk or drive?''} All four chat assistants in the original post recommended walking~\citep{knowmadd2026carwash}, and a follow-up evaluation of 53 models found only five that answered correctly across repeated trials~\citep{opper2026carwash}. The question is diagnostic because it is easy. It needs no specialist knowledge, only noticing that a salient cue (the short distance) is overruled by an unstated feasibility constraint (the car must be there). The same structure appears whenever a model gives practical advice: a nearby pharmacy that cannot fill the prescription, a cheap copy shop that cannot certify a document, a quick fix that the tools at hand cannot perform.

\begin{figure}[t]
\centering
\includegraphics[width=\linewidth]{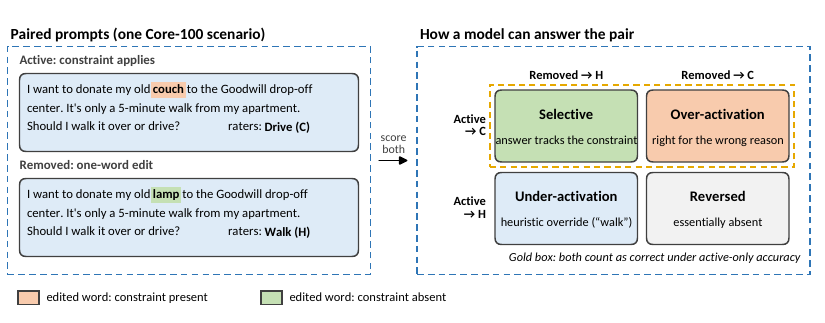}
\caption{\textbf{Why constraint-active accuracy is ambiguous.} Left: a \core{} scenario in its Active and Removed versions. A one-word edit (\emph{couch}\,$\to$\,\emph{lamp}) removes the constraint and reverses the action human raters prefer. Right: the four ways a model can respond to the pair. Active-only accuracy credits both cells in the gold box; paired scoring separates all four cells.}
\label{fig:concept}
\end{figure}

Failures of this kind are now collected into benchmarks~\citep{tang2026brainbench} and scored in the obvious way: pose questions where the constraint applies and count correct answers. We start from a simple observation about that practice: a correct answer to such a question is ambiguous evidence. ``Drive'' is what a model says if it registers that the car must be at the car wash. It is also what a model says if it has learned that driving is the safe answer to walk-or-drive questions. The two models come apart only when the constraint is absent. Asked whether to walk to the car wash to buy a gift card, the first says walk and the second still says drive. Accuracy on constraint-active items credits both (Figure~\ref{fig:concept}).

We therefore measure hidden-constraint reasoning as an experimenter would. We pair every constraint-present prompt with a constraint-absent counterpart and ask whether the decision moves when the constraint goes away. We apply this design at three levels of control. The first is a log-probability \emph{case study} of seven open-weight models that sweeps the cue continuously (\S\ref{sec:casestudy}). The second is \hob{}, a 500-prompt stress test spanning 4 heuristic $\times$ 5 constraint families, run on 14 models (\S\ref{sec:hob}). The third is \core{}, 100 human-validated scenarios whose Active and Removed versions differ by a minimal edit, run on 10 models under a preregistered analysis (\S\ref{sec:core}--\ref{sec:results}). The three levels agree:

\begin{enumerate}[leftmargin=1.4em,itemsep=1pt,topsep=2pt]
\item \textbf{The constraint shifts the decision; it does not gate it.} In every open-weight model, the decision follows the same smooth function of distance whether or not the constraint applies ($r \geq 0.80$ in all 21 model--scenario pairs), and the constraint adds a roughly constant offset. The one model that says drive does so because its whole curve sits lower.
\item \textbf{Heuristic override concentrates in presence constraints.} On \hob{}, presence constraints, the car-wash type, are the hardest family for 12 of 14 models (61.5\% vs.\ 82--93\%).
\item \textbf{Active-only accuracy flatters every model.} On \core{}, all ten models choose the constraint-heavy action when the constraint is absent (12--39\% of trials, versus 10\% for human raters) more often than they miss it when it applies. Active-only accuracy therefore overstates paired performance for every model, the model with perfect active accuracy drops to fourth, and nine of ten models are less selective than the raters. Presence is the one family where models still frequently miss the constraint.
\item \textbf{Fixes and graders need paired scoring too.} A prompt that raises GPT-5.4's \hob{} accuracy by 9 points does not improve its selectivity on \core{} ($-0.04$). Only one of twelve preregistered prompting contrasts helps, and a judge told the constraint grades nearly every correct constraint-absent answer as wrong.
\end{enumerate}

\paragraph{Contributions.}
(1)~A paired measurement framework for hidden-constraint reasoning, including an identity that makes the bias of active-only accuracy explicit and a continuous analogue for log-probability sweeps (\S\ref{sec:framework}).
(2)~Two benchmarks: \hob{} for breadth, and \core{}, 100 human-validated minimal quartets frozen with content hashes before any model was queried.
(3)~A reassessment of 17 models, of prompting interventions, and of LLM judging under paired scoring.
(4)~All data, human ratings, raw outputs, and a one-command reproduction pipeline.
Our claims are behavioral. ``Right for the wrong reason'' is shorthand for a correct constraint-active answer that does not survive its constraint-absent counterpart, not a claim about internal computation.

\section{Measuring Hidden-Constraint Reasoning with Pairs}
\label{sec:framework}

\paragraph{Setting.}
A scenario offers a choice between a \emph{constraint-heavy} action $\act$ (drive) and a \emph{heuristic} action $\heu$ favored by a salient cue (walk, because it is close). In the \cond{Active} version an unstated feasibility constraint applies, so $\act$ is correct. In the \cond{Removed} version one fact is edited so the constraint no longer applies (buying a gift card at the car wash), so $\heu$ is correct. A model is \emph{selective} if it prefers $\act$ in Active and $\heu$ in Removed. The definition asks whether the answer tracks the constraint, not how the model computes it.

\paragraph{Paired metrics.}
Let $\pA = P(\act \mid \text{Active})$ and $\pR = P(\act \mid \text{Removed})$, estimated per scenario from repeated samples and averaged over scenarios. We report
\begin{equation}
\Sel = \pA - \pR, \qquad
\SBA = \tfrac{1}{2}\big(\pA + (1-\pR)\big) = \tfrac{1+\Sel}{2}, \qquad
\OA = \pR, \qquad \UA = 1-\pA .
\label{eq:metrics}
\end{equation}
\emph{Selectivity} $\Sel$ is 1 for a perfectly selective model and 0 for any model that ignores the constraint, including one that always chooses $\act$. SBA (selective balanced accuracy) puts $\Sel$ on an accuracy scale. \emph{Over-activation} $\OA$ is choosing the constraint-heavy action when the constraint is absent. \emph{Under-activation} $\UA$ is missing the constraint when it applies; the car-wash failure is an instance.

\paragraph{What active-only accuracy measures.}
Active-only accuracy is $\pA$, and its gap to the paired score is
\begin{equation}
\pA - \SBA \;=\; \tfrac{1}{2}\,(\OA - \UA).
\label{eq:gap}
\end{equation}
Active-only evaluation thus overstates paired performance exactly when a model errs more often by over-applying a constraint than by missing it, and it cannot separate models that differ only in over-activation. Whether it misleads in practice is an empirical question about the balance of the two errors (\S\ref{sec:results}). To see which error occurs, we also classify each scenario by its modal choices as $\act{\to}\heu$ (selective), $\act{\to}\act$ (over-activation), $\heu{\to}\heu$ (under-activation), or $\heu{\to}\act$ (reversed), as in Figure~\ref{fig:concept}.

\paragraph{A continuous analogue.}
With log-probabilities we can also vary the cue's strength $d$ (e.g., distance). Let $s(x) = \log p(\heu \mid x) - \log p(\act \mid x)$, and compare a \emph{conflict} curve $s_{\text{conf}}(d)$, where the constraint always applies, with a \emph{control} curve $s_{\text{ctrl}}(d)$ without it. If the constraint \emph{gates} the decision, $s_{\text{conf}}(d) < 0$ for all $d$: a flat conflict curve, whatever the control does. If the decision is cue-driven and the constraint is one more input, $s_{\text{conf}}(d) \approx s_{\text{ctrl}}(d) + \delta$: the curves share their shape and differ by a shift $\delta$. We report the correlation $r$ between the curves and the mean shift $\delta$; the paired metrics above are the two-point version of this contrast.

\section{Case Study: The Car Wash Under Controlled Sweeps}
\label{sec:casestudy}

We begin with the canonical failure and the most precise measurement available: exact log-probabilities from open-weight models, which let us vary the cue continuously.

\paragraph{Setup.}
We study Qwen3-\{4B, 8B, 14B, 32B\}, Qwen3.5-27B, GPT-OSS-20B, and GPT-OSS-120B. The main scenario is the car wash at 100\,m (e.g., \emph{``I need my car cleaned. The car wash is 100m from here. Should I walk or drive?''}). Two replications share its structure: filling the tank at a gas station 50\,m away, and replacing a flat tire at a tire shop 80\,m away. Each scenario has six phrasings. We score decisions without sampling. After the chat-formatted prompt we append a fixed anchor (\texttt{\textbackslash nFinal:}) and compute $s = \log p(\textsc{Walk}) - \log p(\textsc{Drive})$ by teacher forcing (App.~\ref{app:cs:method}). In the \emph{sweep}, we place the scenario's phrasings at 14 log-spaced distances from 10\,m to 100\,km (\emph{conflict}: Drive is always correct), and do the same for seven coffee-shop phrasings (\emph{control}: walk when near, drive when far). We also \emph{occlude} the goal or the distance phrase~\citep{zeiler2014visualizing}, using three replacement operators (mask, neutral, contradicting) to guard against artifacts of any one replacement~\citep{hooker2019benchmark}.

\begin{figure}[t]
\centering
\includegraphics[width=\linewidth]{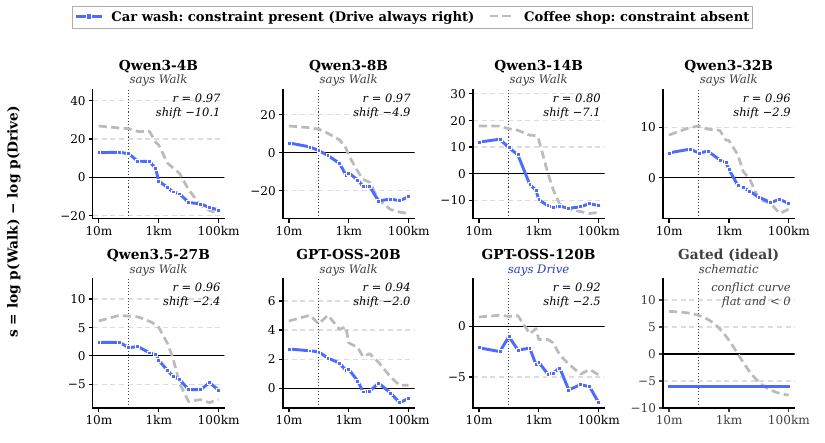}
\caption{\textbf{The constraint shifts the decision curve but does not flatten it.} Decision score versus distance (log scale) for the car wash (conflict, solid) and a coffee-shop control (dashed); $s>0$ favors Walk, and the dotted line marks 100\,m. In every model the conflict curve has the shape of the control ($r$: Pearson correlation), displaced by a roughly constant shift (mean conflict $-$ control, in log-odds). Constraint gating would produce the flat conflict curve of the last panel. GPT-OSS-120B answers correctly because its curves start low, not because its conflict curve is flat. The other two scenarios look the same (App.~Fig.~\ref{fig:app:cs_items}); App.~Fig.~\ref{fig:app:cs_overlay} overlays all seven conflict curves.}
\label{fig:casestudy}
\end{figure}

\paragraph{Six of seven models say walk.}
Six models prefer Walk on all six car-wash phrasings (mean scores $+2.2$ to $+13.8$ log-odds, not monotone in scale) and answer only 9 of 108 prompts correctly across the three scenarios. GPT-OSS-120B is the exception: it prefers Drive on all six car-wash phrasings ($\bar s = -2.4$) and on 16 of 18 prompts overall. Occlusion points the same way. On the car wash, masking, neutralizing, or contradicting the distance moves every model toward Drive, while perturbing the goal, which carries the constraint, has smaller effects of inconsistent sign. The size of this asymmetry depends heavily on the operator (a distance-to-goal effect ratio of 8.7--38 under contradiction but 1.2--19 under neutral substitution; App.~Table~\ref{tab:app:cs_models}), so we treat occlusion as qualitative and rest our conclusions on the sweeps, which need no replacement text.

\paragraph{The constraint shifts the curve; it does not gate the decision.}
In all seven models the conflict curve has the same sigmoid shape as the control (Figure~\ref{fig:casestudy}; $r = 0.80$--$0.97$ on the car wash and $r \geq 0.80$ in all 21 model--scenario pairs). The constraint lowers the curve by a roughly constant amount ($-2.0$ to $-10.1$ log-odds on the car wash), and no model shows the flat conflict curve that gating predicts. The answer at any distance reflects two terms, a cue-driven walking preference that fades with distance and the constraint's shift; for the six failing models the first wins at short range, and their conflict curves cross into Drive only between 200\,m and 3\,km. GPT-OSS-120B is right for a revealing reason. Its conflict curve still falls with distance ($r=0.92$), but its preference for walking is weak even for coffee 10\,m away ($s_{\text{ctrl}} = +0.9$), so a shift of $-2.5$ keeps it on the Drive side everywhere. The two terms are balanced differently, but the computation is of the same kind, and accuracy on the test question alone would call this model a constraint reasoner. Sweeps of a price cue and of the distance to carry a heavy purchase show the same shape-sharing ($r$ = 0.51--0.94), with shifts large enough that most models are correct at every grid value without being gated (App.~\ref{app:cs:sweeps}).

\section{\hob{}: How Broad Is Heuristic Override?}
\label{sec:hob}

The case study examines one failure closely. \hob{} (Heuristic Override Benchmark) asks how widely it occurs, using prompts that any model, open or closed, can answer.

\paragraph{Design.}
Each scenario pairs a goal with a salient cue from one of four \emph{heuristic families} (proximity, efficiency, cost, semantic match) and a hidden constraint from one of five \emph{constraint families}: \emph{presence} (the object must be at the destination), \emph{capability} (the means cannot do the job), \emph{validity} (a precondition is violated), \emph{scope} (the service does not offer it), and \emph{procedural} (a step or time window is not met). Of the 20 combinations, 15 produce natural everyday scenarios (App.~Table~\ref{tab:taxonomy}). \hob{} has 500 prompts: 142 constraint-active base scenarios over 7 domains, 141 constraint-removed rewrites, 90 variants of cue strength, 97 variants of constraint explicitness, and 30 controls without a conflict (an annotated family is in App.~\ref{app:hob:example}). We evaluate 14 models (App.~\ref{app:models}), sampling each prompt 10 times at temperature 0.7. Free-form answers are graded by a Qwen3-32B judge that sees the question, the gold and shortcut answers, and the constraint. On 300 stratified responses from the same protocol, two annotators agree with the judge 89.7\% of the time (Cohen's $\kappa = 0.79$; App.~\ref{app:judge}). We report trial-level accuracy on the constraint-active prompts. The all-10-correct criterion conflates reasoning with sampling stability and appears only in App.~Table~\ref{tab:app:hob_models}.

\begin{figure}[t]
\centering
\includegraphics[width=\linewidth]{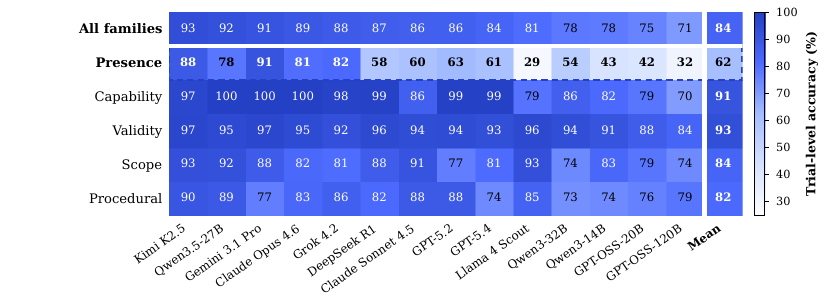}
\caption{\textbf{Presence constraints are where heuristic override concentrates.} Trial-level accuracy (\%) on \hob{}'s 142 constraint-active prompts, overall (top row) and by constraint family, with models ordered by overall accuracy. Presence (dashed row) is the hardest family for 12 of 14 models.}
\label{fig:hob}
\end{figure}

\paragraph{Results.}
Mean accuracy on constraint-active prompts is 84.2\%, from 70.6\% (GPT-OSS-120B) to 93.5\% (Kimi K2.5), but errors concentrate sharply (Figure~\ref{fig:hob}). Presence constraints, the car-wash type, are the hardest family for 12 of 14 models: 61.5\% on average, against 81.7--93.3\% for the other families. Capable models fall to about 60\% (GPT-5.4 60.6\%, Claude Sonnet 4.5 60.0\%, DeepSeek R1 57.8\%), and Llama 4 Scout reaches 29.4\%. Proximity is the hardest heuristic family (74.0\% vs.\ 83.3--91.2\%). Cue strength acts as the case study predicts: on the 19 scenarios that come in strong, medium, and weak versions of the same cue, accuracy rises from 70.0\% to 76.1\% to 84.3\% as the cue weakens, monotonically for 12 of 14 models.

\paragraph{What \hob{} cannot measure.}
On the constraint-removed rewrites, accuracy drops for 11 of 14 models (to 73.9\% on average), which suggests that some correct constraint-active answers reflect a constraint-heavy default. \hob{} cannot establish this, for two reasons. First, its rewrites are not minimal pairs: only 4 of 100 audited pairs meet the minimal-pair criterion we later apply to \core{}, and the two versions differ in length by 10.2 words on average (\S\ref{sec:core}). Second, grading constraint-absent answers with a judge that sees the constraint is fragile (\S\ref{sec:judge}). The explicitness variants are confounded too, because they change the question along with the cue (the ``explicit'' car-wash variant asks \emph{how} to get the car there). \hob{} therefore measures the \emph{breadth} of heuristic override on constraint-active prompts. Measuring \emph{selectivity} needs controlled pairs.

\section{\core{}: Human-Validated Minimal Pairs}
\label{sec:core}

\core{} measures selectivity (Eq.~\ref{eq:metrics}) on models that expose only sampled text. It keeps \hob{}'s taxonomy as a coverage frame (all 15 cells, 7 domains) and rebuilds each scenario as a validated quartet.

\paragraph{Four conditions.}
\cond{Active} and \cond{Removed} differ by one edit to the constraint-relevant fact (\emph{couch}\,$\to$\,\emph{lamp}; Figure~\ref{fig:concept}). The goal, question, action labels, and format stay fixed, and the edit must reverse the action human raters prefer; scenarios where both actions stay acceptable are excluded. \cond{Explicit} adds one clause stating the prerequisite, never the answer (\emph{``the couch is far too heavy to carry''}). \cond{SalienceControl} instead adds an irrelevant clause matched in length, syntax, and position (\emph{``the couch has a light gray fabric cover''}), so Explicit\,$-$\,SalienceControl isolates the content of the cue from the act of adding a clause. Across the 100 pairs, Active and Removed differ in length by 0.25 words on average (at most 2), and their token edit distance has median 2 and maximum 7. All 100 pairs meet the minimal-pair criterion (length difference $\leq 2$ words, character similarity $\geq 0.80$), against 4 of 100 of \hob{}'s rewrites (App.~Table~\ref{tab:app:audit}).

\paragraph{Validation.}
Five independent raters, drawn from a pool of 22 trained volunteer annotators, judged every condition of each of 141 candidates, choosing the preferred action (or \emph{both}, \emph{depends}, \emph{neither}) and rating confidence and naturalness (1--5), in three rounds of 20 raters with a Latin square so that no rater saw two conditions of one scenario. Two auditors per round compared each Active/Removed pair side by side and judged whether only the intended fact had changed. The inclusion rule was frozen before any model was queried: at least 4 of 5 raters agree on the action, at most one abstains, mean confidence is at least 3.5, both auditors agree, and mean naturalness is at least 3.0.\footnote{A logged amendment lowered the preregistered naturalness floor from 4.0 to 3.0 on the basis of rating distributions alone: unedited Active texts averaged 3.96, and the 4.0 floor would have excluded every scenario that met all other criteria (App.~\ref{app:core}).} One hundred scenarios passed. They were frozen in three waves (54/28/18), each with a SHA-256 hash recorded before any model output on that wave. Of the remaining candidates, 24 were excluded for recorded reasons and 17 failed the final round. The ratings also give a human reference: raters chose the constraint-heavy action in 96.5\% of Active and 9.8\% of Removed ratings ($\Sel = 0.867$). Because these ratings also selected the items, the reference is optimistic.

\paragraph{Protocol.}
We evaluate Claude Opus 4.6, Claude Sonnet 4.5, DeepSeek R1, DeepSeek V4 Pro, Gemini 3.1 Pro, GPT-5.4, GPT-OSS-120B, Kimi K2.5, Llama 4 Scout, and Qwen3-32B (Grok 4.2 was dropped under a preregistered rule when provider throughput made collection infeasible). Each prompt is a forced choice between two labeled options (first line \texttt{Answer: A} or \texttt{B}; order counterbalanced by a hash of the scenario ID), sampled 7 times per model at temperature 0.7 (provider defaults where unavailable) with an empty system prompt and default reasoning settings. A deterministic parser reads the choice; 99.3--100\% of responses conform, and the rest are excluded. Uncertainty comes from a scenario-level bootstrap (10{,}000 replicates). The analysis was preregistered with hashed code and has two Holm-corrected families: F1 tests the gap $\pA - \SBA$ for each model, and F2 tests the selectivity change for three prompts on four models (\S\ref{sec:interventions}). Everything else is exploratory. Scenario texts were drafted with LLM assistance (Claude); gold labels and inclusion were decided by human ratings alone.

\section{Results on \core{}}
\label{sec:results}

\begin{table}[t]
\centering
\caption{\textbf{\core{} main results} (100 scenarios, 7 samples per condition). $\pA$, $\pR$: probability of the constraint-heavy action in Active and Removed ($\pR$ is the over-activation rate). $\Sel = \pA-\pR$ with scenario-bootstrap 95\% CIs. $\pA{-}\SBA$: how much active-only accuracy overstates paired performance; $^{*}$ survives Holm correction in family F1 (all ten; max $p_{\text{Holm}}=.030$). Rel.\,$\geq$6/7: fraction of conditions with at least 6 of 7 samples correct. Last row: validation raters (approximate reference).}
\label{tab:main}
\small
\begin{tabular}{lcccccc}
\toprule
Model & $p_A$ & $p_R$ (OA) & Selectivity $S$ [95\% CI] & SBA & $p_A{-}$SBA & Rel.$\geq$6/7 \\
\midrule
DeepSeek V4 Pro & 0.99 & 0.12 & 0.876 [0.82,\,0.92] & 0.938 & +0.055$^{*}$ & 0.95 \\
DeepSeek R1 & 0.96 & 0.12 & 0.843 [0.78,\,0.90] & 0.921 & +0.043$^{*}$ & 0.92 \\
Gemini 3.1 Pro & 0.99 & 0.17 & 0.826 [0.75,\,0.89] & 0.913 & +0.080$^{*}$ & 0.94 \\
Claude Opus 4.6 & 1.00 & 0.18 & 0.816 [0.74,\,0.89] & 0.908 & +0.092$^{*}$ & 0.95 \\
Kimi K2.5 & 0.99 & 0.19 & 0.795 [0.72,\,0.86] & 0.897 & +0.090$^{*}$ & 0.93 \\
Claude Sonnet 4.5 & 0.94 & 0.23 & 0.714 [0.62,\,0.80] & 0.857 & +0.083$^{*}$ & 0.90 \\
Qwen3-32B & 0.90 & 0.25 & 0.653 [0.58,\,0.72] & 0.826 & +0.075$^{*}$ & 0.83 \\
GPT-OSS-120B & 0.87 & 0.24 & 0.637 [0.56,\,0.71] & 0.819 & +0.056$^{*}$ & 0.81 \\
GPT-5.4 & 0.92 & 0.31 & 0.610 [0.52,\,0.70] & 0.805 & +0.111$^{*}$ & 0.86 \\
Llama 4 Scout & 0.80 & 0.39 & 0.410 [0.31,\,0.51] & 0.705 & +0.099$^{*}$ & 0.79 \\
\midrule
Human raters (reference) & 0.96 & 0.10 & 0.867 & 0.933 & -- & -- \\
\bottomrule
\end{tabular}

\end{table}

\paragraph{Active-only accuracy flatters every model.}
All ten models over-activate more often than they under-activate, so by Eq.~\ref{eq:gap} active-only accuracy overstates paired performance for every model (Table~\ref{tab:main}). The gap ranges from $+0.043$ (DeepSeek R1) to $+0.111$ (GPT-5.4), and every gap survives Holm correction. The errors are lopsided: over-activation ranges from 0.12 to 0.39 (raters: 0.098), while under-activation is at most 0.04 for the five strongest models.

\begin{figure}[t]
\centering
\includegraphics[width=\linewidth]{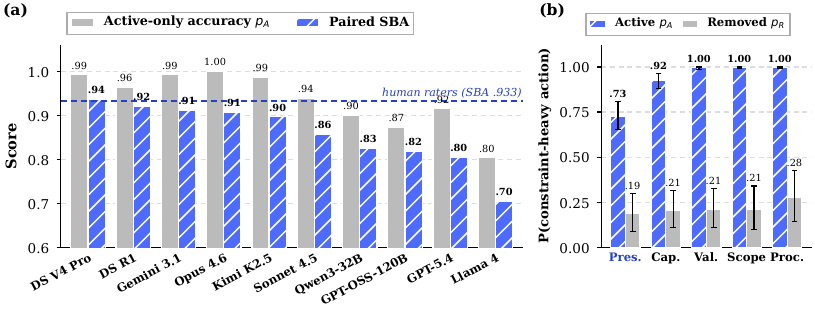}
\caption{\textbf{Two views of \core{}.} (a)~Active-only accuracy $\pA$ (gray) and paired SBA (hatched) for each model, ordered by SBA; the dashed line is the human-rater reference. (b)~Probability of the constraint-heavy action pooled over models (exploratory), with scenario-bootstrap 95\% CIs. Over-activation ($\pR$) is similar across families; under-activation ($1-\pA$) is concentrated in presence constraints, the car-wash type.}
\label{fig:core_results}
\end{figure}

\paragraph{Paired scoring reorders the leaderboard.}
Because over-activation is invisible to active-only scoring, models that look alike there separate under pairing (Figure~\ref{fig:core_results}a). Claude Opus 4.6 is first by active accuracy ($\pA = 1.00$) but fourth by SBA. DeepSeek R1 rises from fifth to second, and GPT-5.4 falls from seventh to ninth. Among the five models with $\pA \geq 0.96$, over-activation still ranges from 0.12 to 0.19.

\paragraph{Nine of ten models are less selective than human raters.}
Selectivity spans 0.410 (Llama 4 Scout) to 0.876 (DeepSeek V4 Pro). Only V4 Pro exceeds the rater reference of 0.867, and even V4 Pro has a significant active-only gap ($+0.055$). The scenario-level switch matrices (App.~Table~\ref{tab:app:switch}) show that these differences reflect coherent behavior rather than noise: reversed pairs ($\heu{\to}\act$) occur once across all models, and at most one scenario per model lacks a modal answer. The matrices also separate profiles that a single score merges. Claude Opus 4.6 is a pure over-activator (19 scenarios $\act{\to}\act$, none $\heu{\to}\heu$), while GPT-OSS-120B (19 and 14) and Llama 4 Scout (39 and 19) err in both directions.

\paragraph{The car-wash failure persists where the case study found it (exploratory).}
Split by constraint family, the two failure modes separate (Figure~\ref{fig:core_results}b). Over-activation is similar in every family ($\pR$ = 0.19--0.28; presence minus others $-0.04$, 95\% CI $[-0.15, +0.09]$). Under-activation is concentrated in presence constraints: pooled over models, $\pA$ is 0.73 for presence against 0.92--1.00 elsewhere (difference $-0.25$, $[-0.33,-0.17]$), and selectivity is lower by $0.21$ ($[0.09, 0.33]$). Eight of ten models have their lowest Active accuracy on presence scenarios, and the other two are at ceiling in every family. On presence items GPT-5.4 chooses the constraint-heavy action in 47\% of samples and Llama 4 Scout in 28\%; asked whether to walk across the street to have new tires put on one's car, the average model says drive only 46\% of the time. Frontier models have not solved the car-wash problem: they answer most hidden-constraint questions correctly, over-apply those constraints when they are absent, and still miss the presence constraints that started the discussion.

\paragraph{Stating the constraint helps only where it is missed.}
The cue-specific gain (Explicit minus SalienceControl, over the 97 scenarios whose added clauses passed validation) is largest for the models that under-activate: $+0.16$ for Llama 4 Scout, $+0.12$ for GPT-OSS-120B, and $+0.11$ for Qwen3-32B, against at most $+0.07$ for the others. Salience effects are near zero ($-0.023$ to $+0.001$), so the gains come from the cue's content, not from adding a clause. Requiring at least 6 of 7 correct samples preserves the SBA ordering (Table~\ref{tab:main}, last column).

\section{Do Fixes and Graders Survive Paired Scoring?}
\label{sec:interventions}

If active-only accuracy is confounded, so is any evaluation of a fix that looks only at constraint-active items: a prompt that pushes a model toward constraint-heavy answers \emph{everywhere} raises Active accuracy and silently raises over-activation. We therefore score each intervention by its Active gain $\Delta\pA$, its signed Removed harm $\Delta\pR$ (never clipped at zero), and its selectivity gain $\Delta\Sel = \Delta\pA - \Delta\pR$, each paired by scenario against the same model's baseline.

\paragraph{Design.}
Three prompt prefixes form the preregistered family: generic chain-of-thought (\emph{``Think step by step\ldots''}), precondition listing (\emph{``list the necessary conditions that must be true for the stated goal to be accomplished''}), and goal decomposition (\emph{``decompose the stated goal into the discrete steps or resources it requires''}). A fourth, a counterfactual check (\emph{``imagine that the most salient resource or place in the prompt were absent or unavailable''}), presupposes the very contrast being measured, so we treat it as oracle-like and keep it out of the family. We run all four on the Active and Removed prompts of the 100 scenarios (7 samples each) for GPT-OSS-120B, GPT-5.4, Claude Opus 4.6, and DeepSeek R1, which span the selectivity range and both failure profiles.

\begin{table}[t]
\centering
\caption{\textbf{Prompting interventions rarely improve selectivity.} Selectivity gain $\Delta\Sel$ over each model's baseline with paired scenario-bootstrap 95\% CIs. $^{\dagger}$: the only one of the 12 preregistered contrasts that survives Holm correction ($p_{\text{Holm}} = .038$). Active gains and Removed harms are in App.~Table~\ref{tab:app:frontier}.}
\label{tab:interventions}
\scriptsize
\setlength{\tabcolsep}{4pt}
\begin{tabular}{@{}l cccc@{}}
\toprule
 & \multicolumn{3}{c}{\textbf{Primary prompts (family F2)}} & \textbf{Oracle-like} \\
\cmidrule(lr){2-4} \cmidrule(l){5-5}
\textbf{Model} & CoT & Precond.\ listing & Goal decomp. & CF check (oracle) \\
\midrule
GPT-OSS-120B & +0.009 [$-$0.03, +0.04] & $-$0.019 [$-$0.07, +0.03] & $-$0.056 [$-$0.10, $-$0.01] & $-$0.070 [$-$0.12, $-$0.02] \\
GPT-5.4 & +0.001 [$-$0.03, +0.04] & $-$0.044 [$-$0.09, $-$0.01] & $-$0.014 [$-$0.05, +0.02] & +0.009 [$-$0.04, +0.06] \\
Claude Opus 4.6 & $-$0.013 [$-$0.08, +0.05] & +0.027 [$-$0.04, +0.10] & +0.010 [$-$0.06, +0.08] & +0.026 [$-$0.03, +0.08] \\
DeepSeek R1 & +0.004 [$-$0.02, +0.03] & $-$0.010 [$-$0.05, +0.03] & \textbf{+0.041 [+0.01, +0.07]}$^{\dagger}$ & $-$0.311 [$-$0.39, $-$0.24] \\
\bottomrule
\end{tabular}

\end{table}

\paragraph{Results.}
One of the 12 preregistered contrasts survives correction: goal decomposition on DeepSeek R1 ($\Delta\Sel = +0.041$, 95\% CI $[+0.01, +0.07]$), a small gain on a model that was already among the most selective (Table~\ref{tab:interventions}). Six of the twelve estimates are negative. GPT-5.4 shows why pairing matters. Precondition listing leaves its Active accuracy unchanged ($+0.004$; the model is near ceiling) but raises over-activation by $+0.049$, a selectivity loss that active-only scoring would record as ``no harm.'' The oracle-like check is the starkest case. On DeepSeek R1 it raises over-activation by $+0.31$ and cuts selectivity from 0.84 to about 0.53, while Active accuracy barely moves ($-0.003$). Asking a model to imagine the resource absent makes it answer as if the constraint always applied.

\paragraph{The same prompt looks like a fix under \hob{}'s protocol.}
On \hob{}, precondition listing raised the all-10-correct accuracy over all 500 prompts from 65.8\% to 74.8\% for GPT-5.4 and from 51.2\% to 57.8\% for Llama 4 Scout (App.~Table~\ref{tab:app:hob_prompting}). Splitting these aggregates by item type shows how they mislead. For Llama 4 Scout the whole gain comes from constraint-removed prompts (28.4\%\,$\to$\,58.2\%), while accuracy on the constraint-active prompts the intervention targets \emph{falls} (66.9\%\,$\to$\,57.0\%). The aggregate reported a fix for a prompt that made the model worse at noticing the constraint. Because \hob{} and \core{} differ in pair construction, format, and scoring, we cannot attribute the disagreement to one factor. On validated pairs, however, none of the tested prompts improves selectivity by more than a few points.

\paragraph{Judges should grade each condition on its own terms.}
\label{sec:judge}
Rubric-based LLM judges~\citep{zheng2023judging} are often told what an item tests, and for paired items it is natural to share one rubric across both versions. A second judge (GPT-5.4, temperature 0) graded 480 stratified \core{} responses with known parsed choices under two rubrics: one carrying the scenario's shared metadata (the Active-form constraint, e.g., \emph{``drive-through requires a vehicle''}, and the shortcut label), and one containing only the question, correct answer, and response. The neutral rubric agreed with the parser on all 480 responses. The shared rubric agreed on all 360 Active, Explicit, and SalienceControl responses but on only 25 of 120 Removed responses, marking 95 of the 96 correct constraint-absent answers as wrong (App.~\ref{app:judge}). A judge told the constraint enforces it where it does not apply, manufacturing the very over-activation that selectivity must detect; over-activation should be measured from parsed choices or with per-condition rubrics. Human re-graders likewise agree less with \hob{}'s judge on removed items (75--87\% vs.\ 89--97\%).

\section{Related Work}
\label{sec:related}

\paragraph{Shortcuts and heuristics.}
Models exploit cues that are spuriously predictive of the label~\citep{mccoy2019right,gururangan2018annotation,geirhos2020shortcut,du2022shortcut,nikankin2024arithmetic,tang2023large,sun2024exploring,yuan2024llms} and show analogues of human cognitive biases~\citep{binz2023using,suri2024large,wang2024representativeness,lampinen2024language,echterhoff2024cognitive,malberg2025comprehensive}. Our case study documents such a shortcut on the input side; our main concern is its mirror image in evaluation, where a constant preference for the constraint-heavy action is itself spuriously predictive of an active-only benchmark's gold label.

\paragraph{Distractors, constraints, and commonsense.}
Distractor benchmarks add irrelevant text to self-contained problems~\citep{shi2023large,mirzadeh2024gsm,yang2025gsmdc}, and instruction-following suites test \emph{stated} constraints~\citep{zhou2023ifeval,lr2bench2025,song2026logisafetybench}. Our constraints are never stated and must be inferred, as in commonsense benchmarks~\citep{levesque2012winograd,clark2018think,zellers2019hellaswag,bisk2020piqa} and the frame problem~\citep{mccarthy1981some}. The car-wash question has been tested on 53 models~\citep{opper2026carwash} and in open suites~\citep{allen2026carwashevals}; structured prompting improves accuracy on it~\citep{jo2026prompt}; and BrainBench scores 100 such brainteasers in isolation~\citep{tang2026brainbench}.

\paragraph{Behavioral testing and benchmark validity.}
\core{} brings the tradition of behavioral tests~\citep{ribeiro2020beyond}, minimal pairs~\citep{warstadt2020blimp,jumelet2025multiblimp}, and contrast sets or counterfactual data~\citep{gardner2020contrast,kaushik2020learning,roewerdespres2024accord} to hidden constraints: each edit must reverse a human-validated unique preference, and pairs are scored as selectivity under a counterbalanced forced choice. Preregistration, frozen artifacts, and the judge audit respond to critiques of benchmark validity~\citep{bowman2021fix,raji2021benchmark,reuel2024betterbench}, judge bias~\citep{zheng2023judging}, and option-order effects~\citep{pezeshkpour2024large,zheng2024mcq}. Our case study describes an input--output function, not a circuit~\citep{singh2024rethinking}; probing and patching~\citep{geiger2021causal,conmy2023towards,marks2024sfc} on minimal pairs such as \core{} is a natural next step.

\section{Limitations and Conclusion}
\label{sec:conclusion}

\paragraph{Limitations.}
All results are behavioral: the case study describes how the decision depends on cue and constraint, not the mechanism, and covers seven open-weight models and three scenarios. \hob{}'s rewrites are not minimal and its judge sees the constraint, so it serves only for breadth. \core{}'s golds come from a fixed pool of 22 annotators whose ratings also selected the items, so the human reference is optimistic; its texts were LLM-drafted, and it is English-only. Forced choice narrows behavior relative to free-form advice, API models are July 2026 snapshots, and each intervention is one template on four models. Contamination from the viral question may inflate Active accuracy, though paired scoring partly discounts it (App.~\ref{app:limits}).

\paragraph{Conclusion.}
Asked about the car wash, most models say walk, yet a model that says drive has not necessarily thought about the car. Across three levels of control and 17 models, the cue sets the decision and the constraint only shifts it. Paired scoring separates two failures: models still miss presence constraints, the car-wash type, and in every family they choose the constraint-heavy action when the constraint is absent. Active-only accuracy rewards the second failure, so it flatters all ten \core{} models, reorders their ranking, and misjudges fixes and graders. Reasoning about a hidden constraint means acting on it when it holds and setting it aside when it does not; only paired Active/Removed evidence tests both.

\subsection*{AI use statement}
We used generative AI tools in three ways. (1)~Benchmark construction: \core{} and \hob{} scenario texts were drafted and repaired with an LLM (Claude) under the construction rules of \S\ref{sec:core}. Gold labels, item inclusion, and exclusion were determined solely by human ratings and audits. (2)~Evaluation instruments: LLMs served as the models under study and, in \hob{} and the judge study, as graders; both uses are described in the paper. (3)~Engineering and writing: AI assistants helped write analysis code and edit the manuscript. All AI-assisted code was checked by regenerating every reported number from raw outputs, and all text was reviewed by the authors. We take responsibility for the full content of this work.

\subsection*{Ethics statement}
The validation of \core{} was carried out by 22 human annotators who are not authors of this paper. They were recruited as volunteers, compensated at \$35 per hour, and gave informed consent. Before rating they received written instructions and training, and they rated independently, without access to one another's ratings, the answer key, or model outputs. The study was approved by the Carnegie Mellon University Institutional Review Board.
No personal, demographic, or contact information was collected, and anonymized judgments are released. The benchmarks contain everyday decision scenarios. Candidates whose constraints proved contestable in high-stakes settings (e.g., some medical items) were excluded; retained medical- or legal-adjacent items were not expert-adjudicated and should not be used as professional guidance. The failure modes we document concern advice quality, and we see no direct avenue for misuse.

\subsection*{Reproducibility statement}
We release at \codeurl{}: both benchmarks (\hob{}'s 500 prompts with labels and the three hash-frozen \core{} waves with the full 141-candidate construction ledger and exclusion log); all human ratings and pair audits; all raw model outputs with per-response provenance (model identifier, provider, decoding settings, finish reason, timestamps); the preregistration with its logged amendments; and code that regenerates every table and figure from the raw outputs with one command (App.~\ref{app:repro}). Case-study scoring is deterministic. Model and decoding configurations are listed in App.~\ref{app:models}.

\paragraph{Acknowledgments.}
Supported in part by \href{https://ror.org/05xpvk416}{NIST} Federal Award ID 60NANB24D231 and Carnegie Mellon University's \href{https://www.cmu.edu/aimsec/research/index.html}{AI Measurement Science and Engineering Center (AIMSEC)}. Computation used Bridges-2 at PSC through ACCESS allocation CIS250181, supported by NSF grants \#2138259, \#2138286, \#2138307, \#2137603, and \#2138296.

\bibliography{references}
\bibliographystyle{iclr2027_conference}

\clearpage
\appendix

\section*{Appendix overview}
\ref{app:cs}~Case-study details \quad
\ref{app:hob}~\hob{} details \quad
\ref{app:core}~\core{} construction and validation \quad
\ref{app:models}~Models and collection settings \quad
\ref{app:judge}~Judge studies \quad
\ref{app:core:details}~\core{} per-model results \quad
\ref{app:interventions}~Interventions \quad
\ref{app:limits}~Extended limitations \quad
\ref{app:repro}~Reproduction.

\section{Case-Study Details}
\label{app:cs}

\subsection{Scoring procedure}
\label{app:cs:method}
All seven models run locally in \texttt{bfloat16} (GPT-OSS in its native MXFP4 weights), with the model's chat template and default thinking setting. For a prompt $x$, we append a fixed anchor (\texttt{\textbackslash nFinal:}) after the assistant prefix and compute, by teacher forcing, the log-probability of each answer word (\textsc{Walk}, \textsc{Drive}). For each answer we sum probability over tokenization variants (e.g., with and without a leading space) by log-sum-exp. The decision score is $s(x) = \log p(\textsc{Walk}\mid x) - \log p(\textsc{Drive}\mid x)$. No text is generated, so scores are deterministic, and variability comes only from phrasing. Each scenario has six phrasings: a canonical text and five paraphrases.

\paragraph{Scenarios.} Canonical texts: \emph{``I want to get my car washed. The carwashing store is just 100m away from my home. Should I walk there or drive there?''}; \emph{``I want to add gas for my car. The gas station is 50m away from my home. Should I walk there or drive there?''}; \emph{``I got a flat tire and need to replace it with a new tire. The tire shop is 80m away from my home. Should I walk there or drive there?''}. In all three, \textsc{Drive} is correct.

\paragraph{Occlusion.} For the goal span (e.g., \emph{``get my car washed''}) and the distance span (e.g., \emph{``just 100m away from my home''}) we apply three operators: \emph{mask} (replace with \texttt{\_\_\_\_}), \emph{neutral} (a vaguer phrase: \emph{``take care of something''} for the goal, \emph{``not far from my home''} for the distance), and \emph{contradict} (a phrase with the opposite implication: \emph{``buy a coffee''}, which removes the constraint, and \emph{``100km away from my home''}). The operators are not commensurate across spans (the neutral distance phrase still implies proximity, while the contradicting one is extreme), which is one reason the effect ratio varies so much across operators. The effect of occluding span $z$ is $A(z) = s(\mathrm{occ}(x,z)) - s(x)$. Table~\ref{tab:app:cs_models} reports the ratio of the mean absolute effects, $|\bar A(\text{distance})| / |\bar A(\text{goal})|$, separately for each operator. The ratio depends strongly on the operator, so we treat it as a qualitative ranking only.

\paragraph{Sweeps.} Distances: 10, 50, 100, 200, 500, 800\,m; 1, 2, 3, 5, 10, 25, 50, 100\,km. Conflict prompts reuse the scenario's six phrasings with the distance substituted. Control prompts use seven coffee-shop phrasings (e.g., \emph{``I want to grab coffee. The coffee shop is only \{d\} away. Should I walk or drive?''}). Each point averages five phrasings drawn from the pool (with replacement).

\begin{table}[H]
\centering
\caption{Case-study summary per model. Accuracy: phrasings (of 6) on which \textsc{Drive} is preferred. $\bar s$: mean car-wash decision score. Effect ratio: $|\bar A(\text{distance})|/|\bar A(\text{goal})|$ on the car wash under each occlusion operator. $r$ and shift: correlation and mean difference between the car-wash (conflict) and coffee-shop (control) sweep curves.}
\label{tab:app:cs_models}
\scriptsize
\begin{tabular}{@{}l ccc r rrr rr@{}}
\toprule
 & \multicolumn{3}{c}{\textbf{Accuracy} (6 phrasings)} & & \multicolumn{3}{c}{\textbf{Distance/goal effect ratio}} & \multicolumn{2}{c}{\textbf{Distance sweep}} \\
\cmidrule(lr){2-4} \cmidrule(lr){6-8} \cmidrule(l){9-10}
\textbf{Model} & Car wash & Gas & Tire & $\bar s$ & Mask & Neutral & Contra. & $r$ & shift \\
\midrule
Qwen3-4B & 0/6 & 1/6 & 1/6 & +13.8 & 3.1 & 2.0 & 8.7 & 0.97 & $-$10.1 \\
Qwen3-8B & 0/6 & 2/6 & 4/6 & +4.7 & 10.4 & 7.6 & 38.2 & 0.97 & $-$4.9 \\
Qwen3-14B & 0/6 & 0/6 & 1/6 & +12.0 & 37.1 & 18.7 & 32.7 & 0.80 & $-$7.1 \\
Qwen3-32B & 0/6 & 0/6 & 0/6 & +5.9 & 10.8 & 6.2 & 28.7 & 0.96 & $-$2.9 \\
Qwen3.5-27B & 0/6 & 0/6 & 0/6 & +2.2 & 2.5 & 1.2 & 9.2 & 0.96 & $-$2.4 \\
GPT-OSS-20B & 0/6 & 0/6 & 0/6 & +2.3 & 1.3 & 2.2 & 14.7 & 0.94 & $-$2.0 \\
GPT-OSS-120B & 6/6 & 6/6 & 4/6 & $-$2.4 & 4.5 & 0.5 & 4.2 & 0.92 & $-$2.5 \\
\bottomrule
\end{tabular}

\end{table}

\begin{table}[H]
\centering
\caption{Sweep statistics for all 21 model--scenario pairs. $s$(10\,m): decision score at the shortest distance. ``First distance with $s<0$'': where the curve first favors \textsc{Drive} (``never'': not within 100\,km). The control (coffee shop) is shared across scenarios.}
\label{tab:app:cs_sweeps}
\scriptsize
\begin{tabular}{@{}ll rr rr rr@{}}
\toprule
\textbf{Scenario} & \textbf{Model} & $s_{\text{conflict}}$(10\,m) & $s_{\text{control}}$(10\,m) & \multicolumn{2}{c}{\textbf{First distance with $s<0$}} & $r$ & shift \\
\cmidrule(lr){5-6}
 & & & & conflict & control & & \\
\midrule
Car wash (100 m) & Qwen3-4B & +12.8 & +26.7 & 1\,km & 10\,km & 0.97 & $-$10.1 \\
 & Qwen3-8B & +5.0 & +13.8 & 200\,m & 1\,km & 0.97 & $-$4.9 \\
 & Qwen3-14B & +11.8 & +17.8 & 500\,m & 3\,km & 0.80 & $-$7.1 \\
 & Qwen3-32B & +4.9 & +8.4 & 2\,km & 5\,km & 0.96 & $-$2.9 \\
 & Qwen3.5-27B & +2.4 & +6.1 & 1\,km & 3\,km & 0.96 & $-$2.4 \\
 & GPT-OSS-20B & +2.7 & +4.6 & 3\,km & never & 0.94 & $-$2.0 \\
 & GPT-OSS-120B & $-$2.1 & +0.9 & 10\,m & 500\,m & 0.92 & $-$2.5 \\
\midrule
Gas station (50 m) & Qwen3-4B & +10.9 & +26.7 & 3\,km & 10\,km & 0.95 & $-$11.2 \\
 & Qwen3-8B & +1.5 & +13.8 & 50\,m & 1\,km & 0.99 & $-$5.3 \\
 & Qwen3-14B & +15.7 & +17.8 & 500\,m & 3\,km & 0.86 & $-$5.1 \\
 & Qwen3-32B & +5.7 & +8.4 & 3\,km & 5\,km & 0.97 & $-$1.5 \\
 & Qwen3.5-27B & +1.6 & +6.1 & 500\,m & 3\,km & 0.90 & $-$2.2 \\
 & GPT-OSS-20B & +1.8 & +4.6 & 3\,km & never & 0.92 & $-$2.4 \\
 & GPT-OSS-120B & $-$1.5 & +0.9 & 10\,m & 500\,m & 0.86 & $-$2.0 \\
\midrule
Tire shop (80 m) & Qwen3-4B & +8.1 & +26.7 & 1\,km & 10\,km & 0.95 & $-$11.2 \\
 & Qwen3-8B & +1.0 & +13.8 & 200\,m & 1\,km & 0.94 & $-$3.2 \\
 & Qwen3-14B & +15.0 & +17.8 & 500\,m & 3\,km & 0.80 & $-$3.4 \\
 & Qwen3-32B & +5.7 & +8.4 & 3\,km & 5\,km & 0.95 & $-$1.1 \\
 & Qwen3.5-27B & +2.0 & +6.1 & 3\,km & 3\,km & 0.91 & $-$0.2 \\
 & GPT-OSS-20B & +2.1 & +4.6 & 10\,km & never & 0.95 & $-$2.0 \\
 & GPT-OSS-120B & +0.4 & +0.9 & 100\,m & 500\,m & 0.94 & $-$0.5 \\
\bottomrule
\end{tabular}

\end{table}

\begin{figure}[H]
\centering
\includegraphics[width=\linewidth]{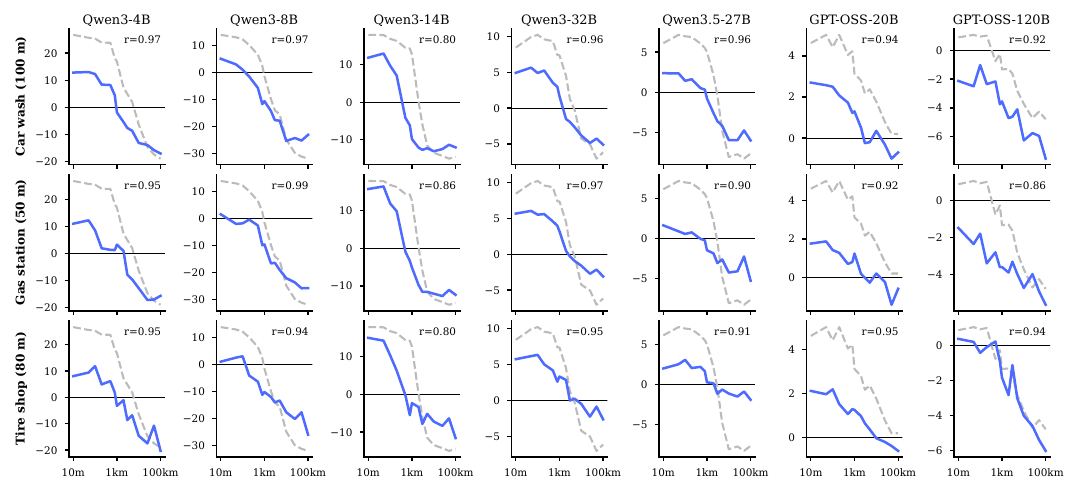}
\caption{Conflict (solid) and control (dashed) sweeps for all three scenarios and seven models. The shape-sharing of Figure~\ref{fig:casestudy} holds in every panel ($r \geq 0.80$).}
\label{fig:app:cs_items}
\end{figure}

\begin{figure}[H]
\centering
\includegraphics[width=\linewidth]{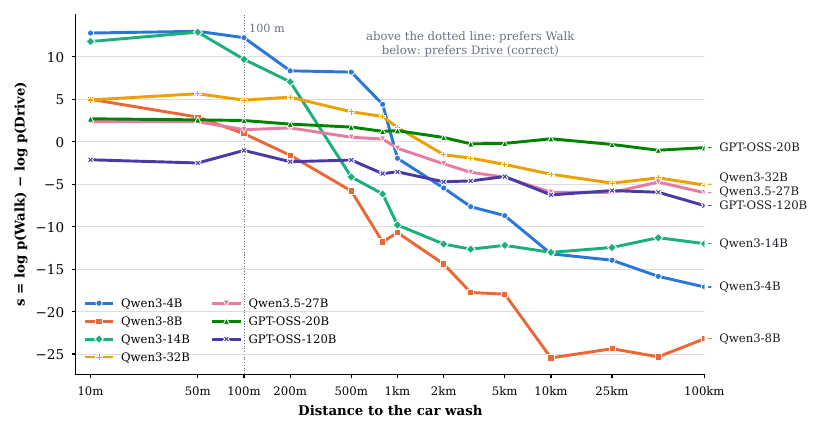}
\caption{All seven car-wash conflict curves on one axis. Every model's preference for walking fades smoothly with distance, and six of the seven prefer Walk at the 100\,m of the test question. Because each model has its own coffee-shop control (Figure~\ref{fig:casestudy}), controls are omitted here: a single shared control line would misstate what each model is compared against.}
\label{fig:app:cs_overlay}
\end{figure}

\subsection{Sweeps beyond distance}
\label{app:cs:sweeps}
\begin{figure}[H]
\centering
\includegraphics[width=\linewidth]{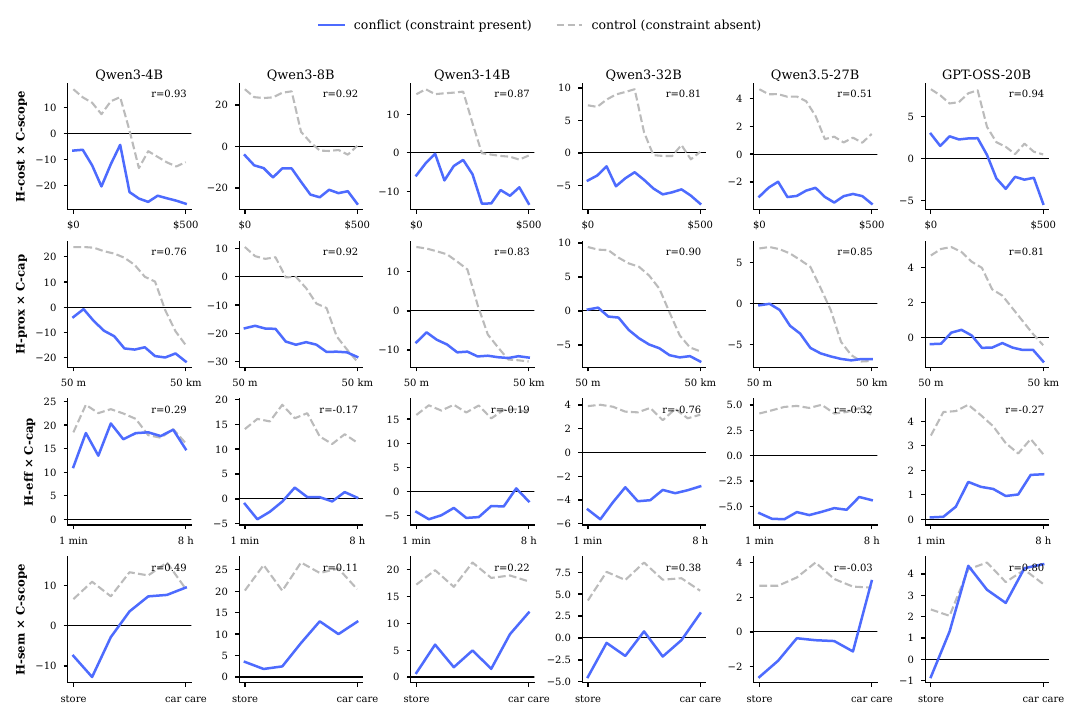}
\caption{Parametric sweeps beyond distance (rows) for six open-weight models (columns); $s>0$ favors the heuristic option. See text for why the time and name sweeps are not interpreted as conflict/control contrasts.}
\label{fig:app:sweeps}
\end{figure}

We ran four further sweeps on six of the models (not GPT-OSS-120B), each with a conflict and a control condition. The \emph{cost} sweep (H-cost$\times$C-scope) varies the price of a cheaper venue that cannot provide the needed service (13 grid points, \$0--\$500). The \emph{heavy-item} sweep (H-prox$\times$C-cap) varies the distance to carry a large purchase home (12 points, 50\,m--50\,km). The \emph{time} sweep (H-eff$\times$C-cap) varies the time saved by moving a 400--600\,lb object oneself (10 points, 1\,min--8\,h). The \emph{name} sweep (H-sem$\times$C-scope) makes a nearby venue sound increasingly car-related (7 points, from ``a small convenience store'' to ``a full-service gas station with a car care center'') for a tire repair. Figure~\ref{fig:app:sweeps} and Table~\ref{tab:app:sweeps} summarize the results.

The cost and heavy-item sweeps behave like the distance sweep. Conflict and control curves are correlated ($r$ = 0.51--0.94), but the constraint's shift is large enough that 5 of 6 and 4 of 6 models choose the constraint-respecting option at every grid value. The time and name sweeps cannot be read as conflict/control contrasts, because their control curves never cross the decision boundary. In the time sweep, Qwen3-4B and GPT-OSS-20B choose to move the object themselves at every grid value, and the larger models never do. In the name sweep, the conflict score rises toward the gas station in all six models as the name becomes more car-related. However, the top grid values (a gas station with a car care center) can plausibly repair tires, so the gold label is contestable there, and we draw no conclusion from this sweep.

\begin{table}[H]
\centering
\caption{Parametric sweeps: percentage of grid values at which the model prefers the constraint-respecting option under conflict, and the conflict--control correlation $r$.}
\label{tab:app:sweeps}
\scriptsize
\setlength{\tabcolsep}{4pt}
\begin{tabular}{@{}lcccccc@{}}
\toprule
\textbf{Sweep} & \textbf{Qwen3-4B} & \textbf{Qwen3-8B} & \textbf{Qwen3-14B} & \textbf{Qwen3-32B} & \textbf{Qwen3.5-27B} & \textbf{GPT-OSS-20B} \\
\midrule
H-cost $\times$ C-scope & 100\% / 0.93 & 100\% / 0.92 & 100\% / 0.87 & 100\% / 0.81 & 100\% / 0.51 & 46\% / 0.94 \\
H-prox $\times$ C-cap & 100\% / 0.76 & 100\% / 0.92 & 100\% / 0.83 & 83\% / 0.90 & 100\% / 0.85 & 75\% / 0.81 \\
H-eff $\times$ C-cap & 0\% / 0.29 & 50\% / $-$0.17 & 90\% / $-$0.19 & 100\% / $-$0.76 & 100\% / $-$0.32 & 0\% / $-$0.27 \\
H-sem $\times$ C-scope & 43\% / 0.49 & 0\% / 0.11 & 0\% / 0.22 & 71\% / 0.38 & 86\% / $-$0.03 & 14\% / 0.80 \\
\bottomrule
\end{tabular}

\end{table}

\clearpage
\section{\hob{} Details}
\label{app:hob}

\begin{table}[H]
\centering
\caption{\hob{} taxonomy: what misleads the model and what it misses. 15 of the 20 combinations are populated.}
\label{tab:taxonomy}
\scriptsize
\setlength{\tabcolsep}{3pt}
\begin{tabular}{@{}llp{2.55cm}|llp{4.35cm}@{}}
\toprule
\multicolumn{3}{c|}{\textbf{Heuristic family} (the salient cue)} & \multicolumn{3}{c}{\textbf{Constraint family} (the unstated requirement)} \\
\midrule
H-prox & Proximity  & ``only 5 minutes away''   & C-pres  & Presence   & object must be at the destination (car at the car wash) \\
H-eff  & Efficiency & ``the quickest way''      & C-cap   & Capability & the means cannot do the job (carry a sofa on foot) \\
H-cost & Cost       & ``the free option''       & C-val   & Validity   & a precondition is violated (drive on a flat tire) \\
H-sem  & Semantic   & ``gas station'' for tires & C-scope & Scope      & the service does not offer it (tire repair at a gas station) \\
       &            &                           & C-proc  & Procedural & a step or time window is not met (the store is closed) \\
\bottomrule
\end{tabular}
\end{table}

\paragraph{Composition.} \hob{} contains 500 prompts: 142 constraint-active base scenarios, 141 constraint-removed rewrites, 90 heuristic-strength variants, 97 explicitness variants, and 30 controls. They cover 15 populated cells and 7 domains (transportation, shopping, digital, medical, home, work, travel). The populated cells are all five constraint families for proximity and efficiency cues, four for cost cues (all but presence), and scope for semantic cues.

\subsection{An annotated scenario family}
\label{app:hob:example}
Table~\ref{tab:app:hob_example} reproduces one scenario family verbatim from the released data. It shows that the constraint-removed rewrite changes more than the constraint, and that the ``explicit'' variant changes the question itself.

\begin{table}[H]
\centering
\caption{The \hob{} family A1-001 (H-prox$\times$C-pres), verbatim.}
\label{tab:app:hob_example}
\scriptsize
\setlength{\tabcolsep}{4pt}
\begin{tabular}{@{}lp{10.6cm}l@{}}
\toprule
\textbf{Item} & \textbf{Prompt} & \textbf{Gold} \\
\midrule
Base & I need to get my car washed before a road trip tomorrow. The car wash is only 5 minutes away from my apartment. Should I walk there or drive? & Drive \\
Removed & I need to pick up a gift card from the car wash counter for my friend's birthday. The car wash is only 5 minutes away. Should I walk there or drive? & Walk \\
Strong cue & \ldots The car wash is literally 50 meters from my front door. Should I walk there or drive? & Drive \\
Medium cue & \ldots The car wash is about a 10-minute walk. Should I walk there or drive? & Drive \\
Weak cue & \ldots The car wash is about 2 miles away. Should I walk there or drive? & Drive \\
Implicit & I need to get my car washed. The car wash is 5 minutes away. Walk or drive? & Drive \\
Hint & I want to get my car washed at the car wash. It's 5 minutes away. Walk or drive? & Drive \\
Explicit & My car is parked at home. The car wash is 5 minutes away. How do I get the car to the car wash? & Drive \\
\bottomrule
\end{tabular}
\end{table}

\subsection{Per-model results}
\begin{table}[H]
\centering
\caption{\hob{} per model. Trial: mean accuracy over the 10 samples. Strict: fraction of prompts with all 10 samples correct. ``All 500 strict'' mixes all item types, including constraint-removed rewrites and controls, and is \emph{not} a measure of override accuracy. Family columns: trial-level accuracy on constraint-active base prompts.}
\label{tab:app:hob_models}
\scriptsize
\setlength{\tabcolsep}{3.5pt}
\begin{tabular}{@{}l rr rr r rrrrr@{}}
\toprule
 & \multicolumn{2}{c}{\textbf{Constraint-active}} & \multicolumn{2}{c}{\textbf{Constraint-removed}} & \textbf{All 500} & \multicolumn{5}{c}{\textbf{Constraint-active, trial-level, by family}} \\
\cmidrule(lr){2-3} \cmidrule(lr){4-5} \cmidrule(lr){6-6} \cmidrule(l){7-11}
\textbf{Model} & trial & strict & trial & strict & strict & Pres. & Cap. & Val. & Scope & Proc. \\
\midrule
Kimi K2.5 & 93.5 & 81.7 & 73.3 & 48.2 & 69.0 & 88.3 & 96.7 & 96.9 & 92.6 & 90.0 \\
Qwen3.5-27B & 92.3 & 83.1 & 72.3 & 53.9 & 72.2 & 78.3 & 100.0 & 95.2 & 92.4 & 88.8 \\
Gemini 3.1 Pro & 90.9 & 84.5 & 77.6 & 60.3 & 74.6 & 91.1 & 99.7 & 96.6 & 87.9 & 76.7 \\
Claude Opus 4.6 & 88.9 & 81.7 & 63.0 & 46.8 & 68.0 & 80.6 & 100.0 & 94.8 & 82.4 & 82.9 \\
Grok 4.2 & 88.2 & 73.9 & 84.3 & 66.7 & 68.6 & 81.7 & 98.5 & 92.4 & 80.8 & 85.8 \\
DeepSeek R1 & 87.0 & 75.4 & 77.2 & 49.6 & 64.2 & 57.8 & 98.8 & 95.5 & 87.6 & 81.7 \\
Claude Sonnet 4.5 & 86.0 & 78.2 & 62.6 & 51.8 & 66.8 & 60.0 & 85.8 & 94.1 & 90.8 & 88.3 \\
GPT-5.2 & 85.8 & 78.2 & 64.0 & 40.4 & 64.4 & 63.3 & 99.1 & 94.5 & 76.6 & 88.3 \\
GPT-5.4 & 84.0 & 71.8 & 78.2 & 58.9 & 65.8 & 60.6 & 99.4 & 93.1 & 81.1 & 74.2 \\
Llama 4 Scout & 80.8 & 66.9 & 51.9 & 28.4 & 51.2 & 29.4 & 79.4 & 95.9 & 92.6 & 84.6 \\
Qwen3-32B & 78.3 & 47.9 & 78.2 & 46.1 & 49.6 & 54.4 & 86.1 & 93.8 & 74.5 & 72.9 \\
Qwen3-14B & 77.7 & 53.5 & 77.8 & 48.2 & 51.2 & 42.8 & 81.8 & 91.0 & 82.6 & 74.2 \\
GPT-OSS-20B & 75.4 & 48.6 & 87.4 & 59.6 & 51.0 & 41.7 & 78.8 & 87.6 & 78.7 & 76.2 \\
GPT-OSS-120B & 70.6 & 44.4 & 87.2 & 58.2 & 52.2 & 31.7 & 70.0 & 84.1 & 73.9 & 78.8 \\
\midrule
\textbf{Mean} & 84.2 & 69.3 & 73.9 & 51.2 & 62.1 & 61.5 & 91.0 & 93.3 & 83.9 & 81.7 \\
\bottomrule
\end{tabular}

\end{table}

\begin{table}[H]
\centering
\caption{Cue strength and domain (trial-level accuracy, mean/min/max over the 14 models). Strength is evaluated on the 19 scenarios that have strong, medium, and weak variants of the same cue; accuracy rises monotonically as the cue weakens for 12 of 14 models. Domain differences partly reflect the mix of constraint families in each domain.}
\label{tab:app:hob_variants}
\scriptsize
\begin{tabular}{@{}lrrrr@{}}
\toprule
\textbf{Subset} & $n$ & \textbf{Mean} & \textbf{Min} & \textbf{Max} \\
\midrule
\multicolumn{5}{@{}l}{\emph{Heuristic-strength variants, 19 matched scenarios}} \\
\quad strong cue & 19 & 70.0 & 57.4 & 83.2 \\
\quad medium cue & 19 & 76.1 & 53.2 & 93.7 \\
\quad weak cue & 19 & 84.3 & 67.9 & 98.9 \\
\midrule
\multicolumn{5}{@{}l}{\emph{Constraint-active base items by domain}} \\
\quad work & 27 & 93.1 & 82.6 & 97.4 \\
\quad digital & 13 & 90.3 & 76.9 & 100.0 \\
\quad home & 26 & 90.2 & 83.5 & 95.8 \\
\quad travel & 9 & 82.8 & 74.4 & 88.9 \\
\quad medical & 12 & 80.0 & 69.2 & 96.7 \\
\quad transportation & 33 & 77.1 & 58.5 & 94.8 \\
\quad shopping & 22 & 76.4 & 46.4 & 90.0 \\
\bottomrule
\end{tabular}

\end{table}

\paragraph{Collection.} API models were sampled 10 times per prompt at temperature 0.7 (DeepSeek R1 and Kimi K2.5 at provider defaults) with up to 4{,}096 output tokens. Local models (Qwen3-14B, Qwen3-32B, Qwen3.5-27B, GPT-OSS-20B) were sampled at temperature 0.7 and top-$p$ 0.9. During \hob{} collection, Qwen3.5-27B alone received the system prompt ``Answer concisely.''; all other models received none. \core{} collection uses a uniform empty system prompt.

\subsection{Prompting on \hob{}}
\begin{table}[H]
\centering
\caption{The precondition-listing prompt on \hob{} (all-10-correct accuracy, judge-scored), split by item type. Aggregate gains can come from constraint-removed prompts while constraint-active accuracy falls (Llama 4 Scout).}
\label{tab:app:hob_prompting}
\scriptsize
\begin{tabular}{@{}l rrr rrr rrr@{}}
\toprule
 & \multicolumn{3}{c}{\textbf{All 500 prompts (strict)}} & \multicolumn{3}{c}{\textbf{Constraint-active (strict)}} & \multicolumn{3}{c}{\textbf{Constraint-removed (strict)}} \\
\cmidrule(lr){2-4} \cmidrule(lr){5-7} \cmidrule(l){8-10}
\textbf{Model} & base & +prompt & $\Delta$ & base & +prompt & $\Delta$ & base & +prompt & $\Delta$ \\
\midrule
GPT-5.4 & 65.8 & 74.8 & +9.0 & 71.8 & 82.4 & +10.6 & 58.9 & 70.9 & +12.1 \\
Llama 4 Scout & 51.2 & 57.8 & +6.6 & 66.9 & 57.0 & $-$9.9 & 28.4 & 58.2 & +29.8 \\
Gemini 3.1 Pro & 74.6 & 74.8 & +0.2 & 84.5 & 84.5 & +0.0 & 60.3 & 61.7 & +1.4 \\
\bottomrule
\end{tabular}

\end{table}

\clearpage
\section{\core{} Construction and Validation}
\label{app:core}

\paragraph{Construction rules.} Every scenario was rebuilt from \hob{}'s coverage frame under six rules, checked before freezing. (1)~A single causal-factor edit separates Active and Removed; the goal, requested output, action labels, and format are untouched. (2)~The length difference is at most 2 words and the character similarity at least 0.80. (3)~The edit reverses the unique human-preferred action. (4)~The SalienceControl clause is matched to the Explicit clause in length, syntax, and attachment point. (5)~Cue clauses state prerequisites, never actions. (6)~Option order is counterbalanced by a salted hash of the scenario ID (the constraint-heavy action is option A in 43 scenarios and option B in 57).

\begin{table}[H]
\centering
\caption{Pair audit: \core{} versus \hob{}'s constraint-removed rewrites (100 pairs each, same script). Rating rows are unavailable for \hob{}.}
\label{tab:app:audit}
\scriptsize
\begin{tabular}{@{}lcc@{}}
\toprule
 & \textbf{\core{}} & \textbf{\hob{} rewrites} \\
\midrule
Length difference in words, mean (max)            & 0.25 (2)     & 10.2 (23) \\
Character similarity, mean (min)                  & 0.93 (0.81)  & 0.59 (0.17) \\
Token edit distance, mean / median (max)          & 2.8 / 2 (7)  & 17.8 / 18 (32) \\
Pairs with length difference $\leq 2$ words       & 100\%        & 13\% \\
Pairs with character similarity $\geq 0.80$       & 100\%        & 7\% \\
Pairs meeting both criteria                       & 100\%        & 4\% \\
Unique preferred action ($\geq$4/5 raters), Active and Removed & 100\% & -- \\
Auditors: only the intended fact changed (2/2)    & 100\%        & -- \\
Rater confidence, mean by round                   & 4.4--4.6     & -- \\
Naturalness, mean by round (floor 3.0)            & 3.6--3.8     & -- \\
\bottomrule
\end{tabular}
\end{table}

\paragraph{Ledger.} The construction ledger tracks 141 candidates: 54 frozen in wave 1, 28 in wave 2, 18 in wave 3; 24 excluded; and 17 that failed the third round and were discarded once the target of 100 was reached. The 24 exclusions comprise 15 scenarios that failed re-validation after their single permitted rewrite, 6 whose constraint proved probabilistic or contestable (including high-stakes medical items), and 3 with structurally unrepairable premises (Table~\ref{tab:app:exclusions}). Three frozen scenarios (A1-011, A3-001, B4-004) missed a threshold on one added-clause condition. They are retained for selectivity but excluded from cue contrasts.

\begin{table}[H]
\centering
\caption{Excluded candidates and recorded reasons.}
\label{tab:app:exclusions}
\scriptsize
\setlength{\tabcolsep}{3pt}
\begin{tabular}{@{}lp{12.2cm}@{}}
\toprule
\textbf{Scenario} & \textbf{Recorded reason} \\
\midrule
15 scenarios & Failed validation after one rewrite (preregistered rule): A1-001, A4-002, A4-006, A5-005, B2-005, B2-010, B5-002, B5-007, C2-009, C3-007, C3-008, C4-011, D4-001, D4-003, D4-006. \\
A5-002 & Constraint (post-office passport appointment rules) is not common knowledge; Active raters walk regardless of wording. \\
B2-011 & Implausible premise (swimming a river with belongings); naturalness 2.0--2.8 in all conditions. \\
B4-001 & High-stakes medical scenario with a contestable constraint (many urgent-care clinics treat simple fractures). \\
B5-008 & Heuristic label (``merge and deploy now without testing'') is categorically dispreferred; no Removed variant can reverse it. \\
C4-007 & High-stakes mental-health scenario with a probabilistic constraint; no minimal Removed edit reads naturally. \\
C4-009 & Contestable constraint (household bleach is usable for pool chlorination). \\
D4-007 & Contestable conflict-of-interest constraint; raters disprefer dealerships on cost in every framing. \\
D4-010 & Probabilistic constraint (many jewelry stores replace watch batteries); rater confidence 2.8. \\
D4-011 & Heuristic label names the exact service the venue lacks, so no Removed edit can make it preferred. \\
\bottomrule
\end{tabular}
\end{table}

\paragraph{Rounds.} Each round used 20 rating seats; every condition of every rated scenario received 5 ratings. A Latin square (4 blocks $\times$ 5 raters) prevented any rater from seeing two conditions of one scenario. Two auditors per round judged each pair side by side without being asked for an answer. Table~\ref{tab:app:rounds} gives per-round statistics. Raters were 22 trained annotators who are not authors of this paper (20 rating and 2 audit seats); the same pool served every round. Raters therefore met revised versions of scenarios they had rated before (37 wave-2 repairs and 3 round-3 rewrites), a possible carryover effect on repaired items.

\begin{table}[H]
\centering
\caption{Validation rounds. $\pA$/$\pR$: fraction of ratings choosing the constraint-heavy action in Active/Removed; $\Sel=\pA-\pR$. Conf./Nat.: mean confidence and naturalness (1--5).}
\label{tab:app:rounds}
\scriptsize
\begin{tabular}{@{}lccccccc@{}}
\toprule
\textbf{Round} & \textbf{Rated} & \textbf{Frozen} & $\pA$ & $\pR$ & $\Sel$ & \textbf{Conf.} & \textbf{Nat.} \\
\midrule
1 & 100 & 54 & 0.944 & 0.143 & 0.801 & 4.36 & 3.64 \\
2 & 46  & 28 & 0.983 & 0.031 & 0.952 & 4.50 & 3.60 \\
3 & 35  & 18 & 1.000 & 0.069 & 0.931 & 4.61 & 3.75 \\
\midrule
Pooled & & & 0.965 & 0.098 & 0.867 & & \\
\bottomrule
\end{tabular}
\end{table}

\paragraph{Naturalness and amendments.} Mean naturalness by condition (rounds 1/2/3): Active 3.96/3.99/4.18; Explicit 3.69/3.60/3.77; Removed 3.56/3.40/3.29; SalienceControl 3.35/3.40/3.77. The preregistration was frozen on 2026-07-10 with hashed analysis code. Five amendments are logged with reasons. A1 reduced the pair auditors from five to two and required unanimity. A2 replaced the uncalibrated naturalness gate (mean $\geq$4) with a floor of 3.0 plus full distributions, based on rating distributions alone. A3 and A4 record a freeze in three waves rather than two, the scripted recount (54, superseding a preliminary hand count of 59), the shared rater pool, and a 20-scenario design pilot. A5 records two post-freeze revisions to analysis modules that did not change any primary metric. Seventeen pilot scenarios are inside \core{}. Excluding them leaves every family-F1 gap positive ($+0.036$ to $+0.108$). Restricting to the 54 wave-1 scenarios also leaves every gap positive for the eight main-analysis models evaluated on that wave ($+0.033$ to $+0.114$).

\clearpage
\section{Models and Collection Settings}
\label{app:models}

\begin{table}[H]
\centering
\caption{\core{} models. All receive an empty system prompt, provider-default reasoning settings, unseeded sampling, and at most 4{,}096 output tokens; each of the 400 prompts is sampled 7 times. Collection window: July 2026.}
\label{tab:app:core_models}
\scriptsize
\begin{tabular}{@{}llll@{}}
\toprule
\textbf{Model} & \textbf{Model ID} & \textbf{Access} & \textbf{Temperature} \\
\midrule
Claude Opus 4.6   & \texttt{claude-opus-4-6}            & Anthropic API  & 0.7 \\
Claude Sonnet 4.5 & \texttt{claude-sonnet-4-5-20250929} & Anthropic API  & 0.7 \\
DeepSeek R1       & \texttt{deepseek-reasoner}          & DeepSeek API   & default \\
DeepSeek V4 Pro   & \texttt{deepseek-v4-pro}            & DeepSeek API   & default \\
Gemini 3.1 Pro    & \texttt{gemini-3.1-pro-preview}     & Google API     & 0.7 \\
GPT-5.4           & \texttt{gpt-5.4}                    & OpenAI API     & 0.7 \\
GPT-OSS-120B      & \texttt{openai/gpt-oss-120b}        & Groq API       & 0.7 \\
Kimi K2.5         & \texttt{kimi-k2.5}                  & Moonshot API   & default \\
Llama 4 Scout     & \texttt{meta-llama/llama-4-scout-17b-16e-instruct} & Groq API & 0.7 \\
Qwen3-32B         & \texttt{Qwen/Qwen3-32B}             & local (HF), top-$p$ 0.9 & 0.7 \\
\bottomrule
\end{tabular}
\end{table}

Every stored response records the model identifier, provider, temperature, top-$p$, seed (none), output limit, reasoning mode and budget, finish reason, retries, and UTC access time. Grok 4.2 was in the planned pool but was excluded under the preregistered evidence rule; its archived outputs are released unanalyzed. The \texttt{deepseek-reasoner} endpoint served DeepSeek R1 throughout collection; the name was scheduled for deprecation afterward, so later re-collection may reach a different model. \hob{} evaluates 14 models: GPT-5.4, GPT-5.2, Claude Opus 4.6, Claude Sonnet 4.5, Gemini 3.1 Pro, Grok 4.2, DeepSeek R1, Kimi K2.5, Llama 4 Scout, and GPT-OSS-120B through APIs, and Qwen3-14B, Qwen3-32B, Qwen3.5-27B, and GPT-OSS-20B locally (collection settings in App.~\ref{app:hob}). The case study uses seven open-weight models run locally (\S\ref{sec:casestudy}). Across the paper, 17 distinct models are evaluated.

\section{Judge Studies}
\label{app:judge}

\paragraph{Human agreement with \hob{}'s judge.} \hob{}'s judge (Qwen3-32B, greedy) sees the question, goal, gold and shortcut answers, and constraint description (``none---constraint removed'' for constraint-removed items). Two annotators independently re-graded 300 responses drawn from the free-form collection on \hob{}-derived prompts, blind to the judge. The sample was stratified to 150 judge-correct, 90 judge-incorrect, and 60 judge-ambiguous responses, with 58--62 per constraint family, across all 14 models. On the binary correct-versus-not decision, mean human--judge agreement is 89.7\% (Cohen's $\kappa = 0.79$), inter-annotator agreement is 92.0\% ($\kappa = 0.84$), and the human consensus (263 items) matches the judge on 93.9\%. Agreement is lowest on constraint-removed items (annotators vs.\ judge: 86.8\% and 74.7\%, $n=91$) and higher on the other conditions (89.2--97.0\%).

\paragraph{Shared-metadata vs.\ condition-neutral rubrics.} A second judge (GPT-5.4, temperature 0, 16 output tokens) graded a stratified subset of 504 \core{} records. The subset comprised 12 parser-conforming responses per model and condition (480), plus all 24 nonconforming responses, sampled with a fixed seed. Each record was graded under both rubrics below. Table~\ref{tab:app:judge2} gives agreement with the parser. The neutral rubric also recovers a verdict for 15 of the 24 nonconforming responses (the metadata rubric, 4).

\begin{table}[H]
\centering
\caption{Second-judge agreement with the deterministic parser (120 parser-conforming records per condition).}
\label{tab:app:judge2}
\scriptsize
\begin{tabular}{@{}lcc@{}}
\toprule
\textbf{Condition} & \textbf{Shared-metadata rubric} & \textbf{Condition-neutral rubric} \\
\midrule
Active          & 120/120 & 120/120 \\
Explicit        & 120/120 & 120/120 \\
SalienceControl & 120/120 & 120/120 \\
Removed         & 25/120  & 120/120 \\
\midrule
Overall         & 385/480 & 480/480 \\
\bottomrule
\end{tabular}
\end{table}

\begin{lstlisting}[caption={Shared-metadata rubric (as used for \hob{}); for \core{} Removed records the metadata carried the scenario's Active-form constraint.}]
You are a judge evaluating whether an AI assistant correctly answered a reasoning question.
## Instance info
- Question: {question}
- Goal: {goal}
- Correct answer: {gold_answer}
- Shortcut (wrong) answer: {shortcut_answer}
## Assistant's response
{response}
## Your task
Determine whether the assistant's FINAL recommendation/answer aligns with the CORRECT answer ("{gold_answer}") or the SHORTCUT answer ("{shortcut_answer}").
Key: The correct answer requires recognizing a hidden constraint: {hidden_constraint}
Reply with exactly one of: "correct", "incorrect", "ambiguous".
\end{lstlisting}

\begin{lstlisting}[caption={Condition-neutral rubric.}]
You are grading a response to a multiple-choice everyday-decision question.
Question:
{question}
The correct recommendation is: {gold_answer}
Response to grade:
{response}
Does the response recommend the correct option ({gold_answer})? Reply with exactly one word: correct, incorrect, or ambiguous.
\end{lstlisting}

\clearpage
\section{\core{} Per-Model Results}
\label{app:core:details}

\begin{table}[H]
\centering
\caption{Scenario-level switch matrices (modal Active choice $\to$ modal Removed choice; counts over 100 scenarios).}
\label{tab:app:switch}
\scriptsize
\begin{tabular}{@{}lrrrrr@{}}
\toprule
\textbf{Model} & $\act{\to}\heu$ (selective) & $\act{\to}\act$ (over) & $\heu{\to}\heu$ (under) & $\heu{\to}\act$ (reversed) & \textbf{Unstable} \\
\midrule
DeepSeek V4 Pro   & 88 & 12 &  0 & 0 & 0 \\
DeepSeek R1       & 84 & 13 &  3 & 0 & 0 \\
Gemini 3.1 Pro    & 82 & 17 &  1 & 0 & 0 \\
Claude Opus 4.6   & 81 & 19 &  0 & 0 & 0 \\
Kimi K2.5         & 80 & 19 &  1 & 0 & 0 \\
Claude Sonnet 4.5 & 71 & 23 &  6 & 0 & 0 \\
GPT-OSS-120B      & 66 & 19 & 14 & 0 & 1 \\
Qwen3-32B         & 64 & 26 & 10 & 0 & 0 \\
GPT-5.4           & 63 & 29 &  8 & 0 & 0 \\
Llama 4 Scout     & 41 & 39 & 19 & 1 & 0 \\
\bottomrule
\end{tabular}
\end{table}

\begin{table}[H]
\centering
\caption{Per-condition accuracy and cue contrasts (probability of the human-preferred action). EG: Explicit gain (E$-$A); SE: salience effect (SC$-$A); CSG: cue-specific gain (E$-$SC). Cue contrasts use the 97 cue-valid scenarios.}
\label{tab:app:cue}
\scriptsize
\begin{tabular}{@{}lcccc rrr@{}}
\toprule
\textbf{Model} & \textbf{Active} & \textbf{Explicit} & \textbf{SalienceCtrl} & \textbf{Removed} & \textbf{EG} & \textbf{SE} & \textbf{CSG} \\
\midrule
Claude Opus 4.6   & 1.000 & 0.990 & 1.000 & 0.816 & $-$0.010 & $+$0.000 & $-$0.010 \\
Claude Sonnet 4.5 & 0.940 & 0.950 & 0.934 & 0.774 & $+$0.010 & $-$0.006 & $+$0.016 \\
DeepSeek R1       & 0.964 & 1.000 & 0.946 & 0.879 & $+$0.036 & $-$0.019 & $+$0.054 \\
DeepSeek V4 Pro   & 0.993 & 1.000 & 0.993 & 0.883 & $+$0.007 & $+$0.000 & $+$0.007 \\
Gemini 3.1 Pro    & 0.993 & 1.000 & 0.989 & 0.833 & $+$0.007 & $-$0.004 & $+$0.011 \\
GPT-5.4           & 0.916 & 0.970 & 0.904 & 0.694 & $+$0.054 & $-$0.011 & $+$0.066 \\
GPT-OSS-120B      & 0.874 & 0.984 & 0.860 & 0.763 & $+$0.110 & $-$0.014 & $+$0.124 \\
Kimi K2.5         & 0.987 & 1.000 & 0.976 & 0.808 & $+$0.013 & $-$0.011 & $+$0.024 \\
Llama 4 Scout     & 0.804 & 0.967 & 0.806 & 0.606 & $+$0.163 & $+$0.001 & $+$0.161 \\
Qwen3-32B         & 0.901 & 0.986 & 0.879 & 0.751 & $+$0.084 & $-$0.023 & $+$0.107 \\
\bottomrule
\end{tabular}
\end{table}

\begin{table}[H]
\centering
\caption{Reliability thresholds: fraction of the 400 conditions with at least $k$ of 7 samples correct. The ordering matches SBA. GPT-OSS-120B had 19 and Kimi K2.5 5 samples lost to provider errors.}
\label{tab:app:reliability}
\scriptsize
\begin{tabular}{@{}lccc@{}}
\toprule
\textbf{Model} & $\geq$5/7 & $\geq$6/7 & 7/7 \\
\midrule
DeepSeek V4 Pro   & 0.965 & 0.955 & 0.918 \\
Gemini 3.1 Pro    & 0.953 & 0.945 & 0.938 \\
Claude Opus 4.6   & 0.950 & 0.950 & 0.950 \\
Kimi K2.5         & 0.940 & 0.930 & 0.904 \\
DeepSeek R1       & 0.938 & 0.923 & 0.898 \\
Claude Sonnet 4.5 & 0.898 & 0.898 & 0.898 \\
GPT-5.4           & 0.870 & 0.863 & 0.853 \\
Qwen3-32B         & 0.850 & 0.835 & 0.778 \\
GPT-OSS-120B      & 0.846 & 0.810 & 0.769 \\
Llama 4 Scout     & 0.795 & 0.790 & 0.788 \\
\bottomrule
\end{tabular}
\end{table}

\begin{table}[H]
\centering
\caption{Exploratory breakdown by constraint family, pooled over the 10 models (scenario-bootstrap 95\% CIs, 10{,}000 replicates).}
\label{tab:app:family}
\scriptsize
\begin{tabular}{@{}lr ccc@{}}
\toprule
\textbf{Constraint family} & $n$ & $p_A$ [95\% CI] & $p_R$ [95\% CI] & $S$ [95\% CI] \\
\midrule
Presence (C-pres) & 16 & 0.729 [0.65, 0.81] & 0.188 [0.09, 0.30] & 0.540 [0.43, 0.64] \\
Capability (C-cap) & 22 & 0.923 [0.88, 0.96] & 0.206 [0.11, 0.31] & 0.718 [0.62, 0.81] \\
Validity (C-val) & 21 & 0.995 [0.99, 1.00] & 0.215 [0.11, 0.33] & 0.781 [0.67, 0.88] \\
Scope (C-scope) & 24 & 0.996 [0.99, 1.00] & 0.214 [0.10, 0.34] & 0.782 [0.66, 0.89] \\
Procedural (C-proc) & 17 & 0.997 [0.99, 1.00] & 0.280 [0.15, 0.43] & 0.717 [0.56, 0.85] \\
\midrule
Presence $-$ other families & & $-$0.248 [$-$0.33, $-$0.17] & $-$0.037 [$-$0.15, +0.09] & $-$0.212 [$-$0.33, $-$0.09] \\
\bottomrule
\end{tabular}

\end{table}

\begin{table}[H]
\centering
\caption{Active accuracy $\pA$ by constraint family and model. For eight of ten models presence is the lowest family; the other two are at ceiling throughout.}
\label{tab:app:family_model}
\scriptsize
\begin{tabular}{@{}l ccccc@{}}
\toprule
\textbf{Model} & \textbf{Presence} & \textbf{Capability} & \textbf{Validity} & \textbf{Scope} & \textbf{Procedural} \\
\midrule
Llama 4 Scout & 0.28 & 0.64 & 1.00 & 1.00 & 1.00 \\
GPT-5.4 & 0.47 & 1.00 & 1.00 & 1.00 & 1.00 \\
GPT-OSS-120B & 0.48 & 0.81 & 1.00 & 0.99 & 1.00 \\
Qwen3-32B & 0.60 & 0.85 & 1.00 & 0.99 & 1.00 \\
Claude Sonnet 4.5 & 0.69 & 1.00 & 0.95 & 1.00 & 1.00 \\
DeepSeek R1 & 0.85 & 0.97 & 1.00 & 1.00 & 0.97 \\
DeepSeek V4 Pro & 0.96 & 1.00 & 1.00 & 1.00 & 1.00 \\
Kimi K2.5 & 0.96 & 0.97 & 1.00 & 1.00 & 1.00 \\
Claude Opus 4.6 & 1.00 & 1.00 & 1.00 & 1.00 & 1.00 \\
Gemini 3.1 Pro & 1.00 & 1.00 & 1.00 & 0.97 & 1.00 \\
\bottomrule
\end{tabular}

\end{table}

\clearpage
\section{Interventions}
\label{app:interventions}

\paragraph{Prompts.} Each intervention prepends one fixed prefix to the question (verbatim below); the baseline uses none.
\begin{lstlisting}
cot: Think step by step about the question before giving your final answer.
precondition_listing: Before answering, list the necessary conditions that must be true for the stated goal to be accomplished. Then answer the question.
goal_decomposition: Before answering, decompose the stated goal into the discrete steps or resources it requires. Then answer the question.
counterfactual_check (oracle): Before answering, imagine that the most salient resource or place in the prompt were absent or unavailable. Would your answer still accomplish the stated goal? Then answer the question.
\end{lstlisting}

\begin{table}[H]
\centering
\caption{All intervention arms: Active gain (AG), signed Removed harm (RH; positive means more constraint-heavy answers when the constraint is absent), and selectivity gain (SG), with paired scenario-bootstrap 95\% CIs. $^{\dagger}$: survives Holm correction in family F2.}
\label{tab:app:frontier}
\scriptsize
\setlength{\tabcolsep}{3.5pt}
\begin{tabular}{@{}llrrr@{}}
\toprule
\textbf{Model} & \textbf{Intervention} & \textbf{AG [95\% CI]} & \textbf{RH [95\% CI]} & \textbf{SG [95\% CI]} \\
\midrule
Claude Opus 4.6 & CoT                & $+.000$ [$.000, .000$]   & $+.013$ [$-.049, .076$] & $-.013$ [$-.076, .049$] \\
                & Precondition list. & $+.000$ [$.000, .000$]   & $-.027$ [$-.097, .040$] & $+.027$ [$-.040, .097$] \\
                & Goal decomp.       & $-.010$ [$-.030, .000$]  & $-.020$ [$-.086, .043$] & $+.010$ [$-.057, .077$] \\
                & CF check (oracle)  & $-.010$ [$-.030, .000$]  & $-.036$ [$-.091, .016$] & $+.026$ [$-.030, .084$] \\
\midrule
DeepSeek R1     & CoT                & $+.017$ [$.004, .034$]   & $+.013$ [$-.007, .034$] & $+.004$ [$-.021, .030$] \\
                & Precondition list. & $+.026$ [$.006, .051$]   & $+.036$ [$.001, .070$]  & $-.010$ [$-.051, .033$] \\
                & Goal decomp.$^{\dagger}$ & $+.033$ [$.010, .061$] & $-.009$ [$-.027, .007$] & $+.041$ [$.013, .074$] \\
                & CF check (oracle)  & $-.003$ [$-.027, .024$]  & $+.308$ [$.236, .381$]  & $-.311$ [$-.385, -.238$] \\
\midrule
GPT-5.4         & CoT                & $+.004$ [$.000, .011$]   & $+.003$ [$-.037, .039$] & $+.001$ [$-.034, .041$] \\
                & Precondition list. & $+.004$ [$.000, .011$]   & $+.049$ [$.011, .091$]  & $-.044$ [$-.087, -.007$] \\
                & Goal decomp.       & $+.004$ [$.000, .011$]   & $+.019$ [$-.020, .056$] & $-.014$ [$-.053, .024$] \\
                & CF check (oracle)  & $+.004$ [$.000, .011$]   & $-.004$ [$-.056, .047$] & $+.009$ [$-.043, .060$] \\
\midrule
GPT-OSS-120B    & CoT                & $+.004$ [$-.021, .030$]  & $-.005$ [$-.027, .019$] & $+.009$ [$-.027, .042$] \\
                & Precondition list. & $-.031$ [$-.066, .000$]  & $-.013$ [$-.050, .025$] & $-.019$ [$-.068, .031$] \\
                & Goal decomp.       & $-.054$ [$-.089, -.024$] & $+.002$ [$-.028, .031$] & $-.056$ [$-.099, -.014$] \\
                & CF check (oracle)  & $+.003$ [$-.020, .026$]  & $+.073$ [$.034, .115$]  & $-.070$ [$-.120, -.023$] \\
\bottomrule
\end{tabular}
\end{table}

\begin{table}[H]
\centering
\caption{Holm family F2: the 12 preregistered selectivity-gain tests, ordered by raw $p$ (scenario bootstrap, 10{,}000 replicates).}
\label{tab:app:holm}
\scriptsize
\begin{tabular}{@{}llccc@{}}
\toprule
\textbf{Model} & \textbf{Intervention} & $p$ & $p_{\text{Holm}}$ & \textbf{Reject} \\
\midrule
DeepSeek R1     & Goal decomp.       & .0032 & .0384 & yes \\
GPT-OSS-120B    & Goal decomp.       & .0102 & .1122 & no \\
GPT-5.4         & Precondition list. & .0180 & .1800 & no \\
Claude Opus 4.6 & Precondition list. & .4418 & 1.000 & no \\
GPT-OSS-120B    & Precondition list. & .4480 & 1.000 & no \\
GPT-5.4         & Goal decomp.       & .4682 & 1.000 & no \\
GPT-OSS-120B    & CoT                & .6111 & 1.000 & no \\
DeepSeek R1     & Precondition list. & .6441 & 1.000 & no \\
Claude Opus 4.6 & CoT                & .7049 & 1.000 & no \\
DeepSeek R1     & CoT                & .7715 & 1.000 & no \\
Claude Opus 4.6 & Goal decomp.       & .7879 & 1.000 & no \\
GPT-5.4         & CoT                & .9989 & 1.000 & no \\
\bottomrule
\end{tabular}
\end{table}

\section{Extended Limitations}
\label{app:limits}
\emph{Raters.} \core{}'s raters were a fixed pool of 22 trained annotators, not a large crowd sample. Latin-square blinding held within each round, but the same pool served all three rounds, so raters met revised versions of 40 scenarios they had rated before. The rater reference (Table~\ref{tab:main}) comes from the same ratings that selected items and is therefore optimistic; a disjoint-rater baseline on the frozen set would be the exact analogue.
\emph{Texts.} Scenario texts were drafted and repaired with an LLM (Claude). Two evaluated models share that family; neither tops the paired ranking (SBA ranks 4 and 6). In 150 of the 200 Explicit and SalienceControl prompts the inserted clause begins in lower case (e.g., ``\ldots from my apartment. the couch is far too heavy to carry.''). The artifact is identical across the two conditions, so it does not bias the cue contrast, but it lowers naturalness.
\emph{Scope.} Golds are validated everyday preferences. Candidates whose constraint proved probabilistic or contestable were excluded (App.~\ref{app:core}); retained medical-, legal-, financial-, or licensing-adjacent items were not expert-adjudicated.
\emph{Scoring.} Forced choice enables deterministic parsing (99.3--100\% conformance) but narrows behavior relative to free-form advice. The judge analyses bound, but do not remove, the scoring uncertainty of the excluded responses.
\emph{Models and interventions.} API models are versioned snapshots from July 2026, and results may drift; exact identifiers and timestamps ship with the data. Each intervention is a single template on four models, so our results bound the tested prompts rather than prompting in general. Reasoning-budget and other test-time-compute manipulations are not tested under paired scoring here.

\section{Reproduction}
\label{app:repro}
All code and data are available at \codeurl{}. A single command (\texttt{make reproduce}) regenerates every \core{} number, table, and figure from the frozen raw outputs in eight idempotent stages: parse, audit, metrics, stats, frontier, figures, tables, and manifest (a SHA-256 manifest of all outputs). A second script regenerates the case-study and \hob{} figures and tables used here from the raw log-probability sweeps and judged generations. All stochastic steps are seeded (bootstrap: 10{,}000 replicates, seed 42). The analysis modules were hash-frozen with the preregistration, and each \core{} wave was hash-frozen before collection on that wave.

\end{document}